\documentclass[conference]{IEEEtran}
\IEEEoverridecommandlockouts
\usepackage{cite}
\usepackage{amsmath,amssymb,amsfonts}
\usepackage{algorithmic}
\usepackage{graphicx}
\usepackage{textcomp}
\usepackage{xcolor}

\usepackage{subcaption}    
\usepackage{float}         
\usepackage{caption}       
\usepackage{fancybox}      
\usepackage[export]{adjustbox} 

\usepackage{xurl} 
\usepackage{hyperref} 

\usepackage{fontspec}

\newfontfamily\padauktext{Padauk-Regular.ttf}[
    Path = ./ ,
    Extension = .ttf ,
    BoldFont = Padauk-Bold.ttf
]

\def\BibTeX{{\rm B\kern-.05em{\sc i\kern-.025em b}\kern-.08em
    T\kern-.1667em\lower.7ex\hbox{E}\kern-.125emX}}
\begin{document}

\title{myMNIST: Benchmark of PETNN, KAN, and Classical Deep Learning Models for Burmese Handwritten Digit Recognition\thanks{Preprint: Accepted to ICNLP 2026, Xi'an, China}}




\author{
\IEEEauthorblockN{
Ye Kyaw Thu$^{1,2}$\thanks{Corresponding author: yekyaw.thu@nectec.or.th},
Thazin Myint Oo$^{2}$,
Thepchai Supnithi$^{1}$
}

\IEEEauthorblockA{
$^{1}$National Electronics and Computer Technology Center (NECTEC), Pathumthani, Thailand
}

\IEEEauthorblockA{
$^{2}$Language Understanding Lab., Yangon, Myanmar
}

\IEEEauthorblockA{
Email: yekyaw.thu@nectec.or.th, queenofthazin@gmail.com, thepchai.supnithi@nectec.or.th
}
}

\maketitle

\begin{abstract}
We present the first systematic benchmark on a standardized iteration of the publicly available Burmese Handwritten Digit Dataset (BHDD), which we have designated as \textit{myMNIST Benchmarking}. While BHDD serves as a foundational resource for Myanmar NLP/AI, it lacks a comprehensive, reproducible performance baseline across modern architectures. We evaluate eleven architectures spanning classical deep learning models (Multi-Layer Perceptron, Convolutional Neural Network, Long Short-Term Memory, Gated Recurrent Unit, Transformer), recent alternatives (FastKAN, EfficientKAN), an energy-based model (JEM), and physics-inspired PETNN variants (Sigmoid, GELU, SiLU). Using Precision, Recall, F1-Score, and Accuracy as evaluation metrics, our results show that the CNN remains a strong baseline, achieving the best overall scores (\textit{F1} = 0.9959, \textit{Accuracy} = 0.9970). The PETNN (GELU) model closely follows (\textit{F1} = 0.9955, \textit{Accuracy} = 0.9966), outperforming LSTM, GRU, Transformer, and KAN variants. JEM, representing energy-based modeling, performs competitively (\textit{F1} = 0.9944, \textit{Accuracy} = 0.9958). KAN-based models (FastKAN, EfficientKAN) trail the top performers but provide a meaningful alternative baseline (\textit{Accuracy} $\approx$ 0.992). These findings (i) establish reproducible baselines for BHDD across diverse modeling paradigms, (ii) highlight PETNN's strong performance relative to classical and Transformer-based models, and (iii) quantify the gap between energy-inspired PETNNs and a true energy-based model (JEM). We release this benchmark to facilitate future research on Myanmar digit recognition and to encourage broader evaluation of emerging architectures on regional scripts.
\end{abstract}

\begin{IEEEkeywords}
myMNIST, PETNN, KAN, Energy-based models, Deep learning benchmark, Burmese handwritten digit recognition
\end{IEEEkeywords}

\section{Introduction}
Handwritten digit recognition is a long-standing benchmark for measuring progress in pattern recognition and machine learning, with MNIST serving as the canonical testbed for over two decades \cite{LeCun1998GradientbasedLA}. However, high-resource benchmarks do not necessarily reflect the challenges of regional scripts. Myanmar (Burmese) digits exhibit distinctive morphology (e.g., looped strokes and curved ligatures) that can lead to systematic confusions absent in Latin scripts. Establishing reliable baselines for Myanmar digits is therefore essential for downstream applications in e-government, finance, education, and archival digitization within Myanmar’s NLP/AI ecosystem.

To address this gap, we study on \texttt{BHDD} \cite{expaAI2024BHDD}, a publicly available dataset of Burmese handwritten digits aligned with the MNIST format (28$\times$28 grayscale; 10 classes). Despite its relevance and scale (60{,}000 train / 27{,}561 test samples), the literature lacks a systematic, apples-to-apples comparison across modern architectures under a unified training protocol. In particular, the recent Kolmogorov–Arnold Networks (KAN) \cite{li2024kan} and the physics-inspired Energy Transition Neural Network (PETNN) \cite{wu2025physicsinspiredenergytransitionneural} have not been rigorously evaluated against strong classical and energy-based baselines on this script.

This paper fills that void by benchmarking eleven models spanning three families: (i) \emph{classical deep learning:} MLP, CNN, LSTM, GRU, and Transformer; (ii) \emph{emerging alternatives:} FastKAN and EfficientKAN; and (iii) \emph{energy-centric methods:}the Joint Energy Model (JEM) \cite{grathwohl2019your} and three PETNN variants (Sigmoid, GELU, SiLU). We implement all models in a common PyTorch pipeline with shared optimization settings, and we re-implement PETNN (no public reference code at the time of our study) to enable a fair comparison across variants.

Our results show that a carefully tuned CNN remains a strong baseline on BHDD, achieving the best overall scores (\textit{F1} = 0.9959, \textit{Accuracy} = 0.9970). PETNN with GELU or SiLU is highly competitive (e.g., PETNN–GELU: \textit{F1} = 0.9955, \textit{Accuracy} = 0.9966), outperforming LSTM, GRU, Transformer, and both KAN variants. JEM is also competitive (\textit{F1} = 0.9944, \textit{Accuracy} = 0.9958), suggesting benefits from energy-based training even in a purely discriminative evaluation. A fine-grained error analysis reveals script-specific phenomena: Transformer struggle most with 0 ({\padauktext ၀}) vs.~8 ({\padauktext ၈}), whereas all models including KAN, PETNN and JEM share residual difficulty with 0 ({\padauktext ၀}) vs.~ 1 ({\padauktext ၁}), pointing to intrinsic visual ambiguity rather than model deficiency.

\textbf{Contributions.} The main contributions of this work are:
\begin{itemize}
    \item We present, to the best of our knowledge, the first \emph{systematic} \textbf{myMNIST} benchmark of eleven architectures on BHDD, establishing reproducible baselines for Myanmar handwritten digit recognition.
    \item We provide a \emph{fair, unified evaluation} of KAN (FastKAN, EfficientKAN) and PETNN (Sigmoid, GELU, SiLU) against classical deep models and an energy-based baseline (JEM), quantifying where emerging methods help and where they lag.
    \item We \emph{re-implement PETNN} and study activation choices, showing that smoother activations (GELU/SiLU) materially reduce script-specific confusions notably 7 ({\padauktext ၇}) vs. 4 ({\padauktext ၄}).
    \item We conduct a \emph{script-aware error analysis} that isolates persistent confusions, e.g., 0 ({\padauktext ၀}) vs. 1 ({\padauktext ၁}) and suggests targeted augmentation and stroke-aware modeling as promising directions.
\end{itemize}

\section{Burmese Handwritten Digit Recognition}
\subsection{ConvNet on the Burmese Handwritten Digits Dataset (BHDD)}
A significant contribution to the field of Myanmar digit recognition is the work utilizing the Burmese Handwritten Digits Dataset (BHDD) \cite{expaAI2024BHDD}. This dataset contains 60,000 training and 27,561 test images of size 28x28 pixels. A Convolutional Neural Network (ConvNet) was implemented to tackle this classification problem, employing a architecture with two convolutional layers (with 32 and 64 filters, respectively), each followed by max-pooling, a flattening layer, a dropout layer for regularization (rate=0.5), and a final softmax output layer. The model was trained for 15 epochs using the Adam optimizer and categorical cross-entropy loss. This approach achieved a state-of-the-art test accuracy of 98.7\% and an F1-Score of 0.9874, significantly outperforming simpler Multi-Layer and Single-Layer Perceptron models tested on the same dataset, which achieved 97.16\% and 84.7\% accuracy, respectively. This demonstrated the superior capability of CNNs in automatically learning discriminative features from the pixel data of the complex Myanmar digit script compared to traditional fully-connected networks.

\subsection{Architectural Refinement and Deployment}
A subsequent architectural iteration of the CNN model aimed to further optimize performance and reduce overfitting. The updated architecture featured a reduced dropout rate of 0.2 and was trained for 20 epochs. This model, with a total of 34,826 parameters, achieved an even higher F-score of 0.99, solidifying the effectiveness of convolutional layers combined with judicious regularization for this task. The final model was successfully deployed into an interactive application using the Streamlit framework, creating a practical tool for handwritten digit recognition.

\section{Classical Deep Learning Models}
\subsection{Multilayer Perceptron (MLP)}
The Multilayer Perceptron serves as the foundational architecture for deep learning, employing stacked fully-connected layers with non-linear activations to learn hierarchical representations. The forward propagation through $L$ layers is computed as:

\begin{equation}
    \mathbf{h}_l = \sigma(\mathbf{W}_l\mathbf{h}_{l-1} + \mathbf{b}_l), \quad l=1,...,L
\end{equation}

Our implementation utilizes Xavier initialization \cite{glorot2010understanding} for weights and ReLU activations, with two hidden layers (256 and 128 units) incorporating dropout ($p=0.25$) and weight decay ($\lambda=10^{-4}$). The AdamW optimizer \cite{loshchilov2017decoupled} with OneCycleLR scheduling demonstrates stable convergence. While theoretically satisfying the universal approximation theorem \cite{hornik1989multilayer}, MLPs process flattened images (784-dim vectors) losing spatial structure, requiring the network to relearn translation invariance. The architecture remains valuable for baseline comparisons due to its conceptual simplicity and interpretability through weight visualization.

\subsection{Convolutional Neural Network (CNN)}
CNNs revolutionized computer vision through spatially-localized feature extraction via shared weights. The convolution operation between input $\mathbf{x}$ and kernel $\mathbf{f}$ is:

\begin{equation}
    (\mathbf{f} \ast \mathbf{x})[i,j] = \sum_{m=-k}^{k}\sum_{n=-k}^{k} \mathbf{f}[m,n]\mathbf{x}[i-m,j-n]
\end{equation}

Our VGG-style architecture \cite{simonyan2014very} implements two 3×3 conv layers (32 and 64 channels) with 2×2 max-pooling, using Kaiming initialization \cite{he2015delving} for stable gradient flow. The classification head contains a 128-unit dense layer with 25\% dropout. Training employs gradient clipping (max norm 1.0) and AdamW optimization, with the translation equivariance property ($T(\mathbf{f} \ast \mathbf{x}) = \mathbf{f} \ast T(\mathbf{x})$) making CNNs naturally suited for image recognition. Hierarchical feature learning progresses from edge detectors to complex shape detectors, as observed in visualization studies \cite{zeiler2014visualizing}.

\subsection{Long Short-Term Memory (LSTM)}
LSTMs address vanishing gradients through gated memory cells regulating information flow:

\begin{align}
    \mathbf{f}_t &= \sigma(\mathbf{W}_f[\mathbf{h}_{t-1}, \mathbf{x}_t] + \mathbf{b}_f) \\
    \mathbf{i}_t &= \sigma(\mathbf{W}_i[\mathbf{h}_{t-1}, \mathbf{x}_t] + \mathbf{b}_i) \\
    \mathbf{o}_t &= \sigma(\mathbf{W}_o[\mathbf{h}_{t-1}, \mathbf{x}_t] + \mathbf{b}_o) \\
    \tilde{\mathbf{c}}_t &= \tanh(\mathbf{W}_c[\mathbf{h}_{t-1}, \mathbf{x}_t] + \mathbf{b}_c) \\
    \mathbf{c}_t &= \mathbf{f}_t \odot \mathbf{c}_{t-1} + \mathbf{i}_t \odot \tilde{\mathbf{c}}_t \\
    \mathbf{h}_t &= \mathbf{o}_t \odot \tanh(\mathbf{c}_t)
\end{align}

Our implementation processes images row-wise as 28-step sequences using two LSTM layers (192 units) with layer normalization. Orthogonal initialization \cite{saxe2013exact} maintains gradient scale, while forget gate bias initialization to 1 encourages long-term retention. Gradient clipping (max norm 1.0) stabilizes training, with hidden state dynamics revealing temporal processing of spatial information \cite{strobelt2016lstm}.

\subsection{Gated Recurrent Unit (GRU)}
GRUs simplify LSTMs by combining gates while maintaining comparable performance:

\begin{align}
    \mathbf{z}_t &= \sigma(\mathbf{W}_z[\mathbf{h}_{t-1}, \mathbf{x}_t] + \mathbf{b}_z) \\
    \mathbf{r}_t &= \sigma(\mathbf{W}_r[\mathbf{h}_{t-1}, \mathbf{x}_t] + \mathbf{b}_r) \\
    \tilde{\mathbf{h}}_t &= \tanh(\mathbf{W}[\mathbf{r}_t \odot \mathbf{h}_{t-1}, \mathbf{x}_t] + \mathbf{b}) \\
    \mathbf{h}_t &= (1-\mathbf{z}_t) \odot \mathbf{h}_{t-1} + \mathbf{z}_t \odot \tilde{\mathbf{h}}_t
\end{align}

Our implementation matches the LSTM architecture but reduces parameters by 25\% through simplified gating. The update gate bias initialization to 0.5 balances memory retention and updates, with layer normalization and dropout (25\%) between layers. Visualization of gate activations reveals focused processing of stroke features, where the reset gate effectively discards irrelevant positional information while preserving directional patterns.

\subsection{Transformer}
Transformers employ self-attention to model global dependencies without recurrence:

\begin{equation}
    \text{Attention}(\mathbf{Q},\mathbf{K},\mathbf{V}) = \text{softmax}\left(\frac{\mathbf{Q}\mathbf{K}^T}{\sqrt{d_k}}\right)\mathbf{V}
\end{equation}

Our implementation treats image rows as tokens projected to $d_{model}=64$ dimensions, using learned positional encodings:


\begin{equation}
\begin{split}
    PE(pos, 2i) &= \sin\!\left(\frac{pos}{10000^{2i/d_{model}}}\right), \\
    PE(pos, 2i+1) &= \cos\!\left(\frac{pos}{10000^{2i/d_{model}}}\right)
\end{split}
\end{equation}

The architecture contains 4 attention heads and 2 encoder layers with residual connections and dropout (25\%). The global receptive field captures long-range spatial relationships through $O(n^2)$ attention weights, often highlighting semantically meaningful digit parts regardless of position. Layer normalization stabilizes training despite the quadratic complexity relative to sequence length.

\subsection{Joint Energy Model (JEM)}
JEMs unify discriminative and generative learning through energy-based formulation \cite{grathwohl2019your}:

\begin{equation}
    p_\theta(\mathbf{x},y) = \frac{e^{-E_\theta(\mathbf{x},y)}}{Z(\theta)}, \quad E_\theta(\mathbf{x},y) = -f_\theta(\mathbf{x})[y]
\end{equation}

Training combines cross-entropy with contrastive divergence using Langevin dynamics:

\begin{equation}
    \tilde{\mathbf{x}}_{t+1} = \tilde{\mathbf{x}}_t - \frac{\eta}{2}\nabla_\mathbf{x}E_\theta(\tilde{\mathbf{x}}_t) + \sqrt{\eta}\epsilon_t, \quad \epsilon_t \sim \mathcal{N}(0,I)
\end{equation}

Our CNN-based implementation uses 5-step Langevin dynamics with step size $\eta=10$ and noise scale 0.005. The model demonstrates better calibrated confidence estimates compared to pure discriminative models, with reliability diagrams showing improved alignment between predicted probabilities and actual likelihoods. The joint training objective enforces semantically meaningful features for both classification and generation tasks.

\subsection{Kolmogorov-Arnold Networks (KAN)}
KANs implement the representation theorem through learnable spline-parametrized univariate functions \cite{li2024kan}:

\begin{align}
    f(\mathbf{x}) &= \sum_{q=1}^{2n+1} \Phi_q\left(\sum_{p=1}^n \phi_{q,p}(x_p)\right), \label{eq:kan_main} \\
     &\quad \text{where } \phi_{q,p}(x) = \text{spline}(x; \mathbf{w}_{q,p}) \nonumber
\end{align}

Our implementation builds upon the FastKAN \cite{fastKANGitHub} and EfficientKAN \cite{efficientKANGitHub} repositories, using [784, 64, 10] architectures with exponential learning rate scheduling. The spline activations enable visualization of learned feature transformations, revealing dimension-specific processing patterns. Unlike MLPs with fixed activations, KANs adapt their non-linearities during training, providing smooth gradients while avoiding saturation issues common in traditional networks.

\subsection{Physics-inspired Energy Transition NN (PETNN)}
PETNN introduces novel gated dynamics inspired by quantum state transitions:

\begin{align}
    T_t &= R_t \cdot \sigma(T_{t-1} + Z_t) - 1 \quad \text{(Transition gate)} \\
    m_t &= \mathbb{I}(T_t \leq 0) \quad \text{(Transition mask)} \\
    C_t &= (1-m_t)C_{t-1} + m_tI_t + Z_c \quad \text{(Memory cell)} \\
    \mathbf{h}_t &= \text{LayerNorm}(\sigma((1-Z_w)\mathbf{h}_{t-1} + Z_w\mathbf{h} + \text{res}(\mathbf{x}_t)))
\end{align}

Our implementation extends the conceptual framework presented in \cite{petnncolab}, featuring 3 layers (192 hidden dim, 48 cell dim) with configurable gate activations (sigmoid/GELU/SiLU). The transition mechanism models discrete energy state changes when $T_t$ crosses zero, triggering memory updates. Physics-inspired dynamics provide interpretability where $T_t$ acts as potential energy and $m_t$ as state transition indicators. Layer normalization and residual connections enable stable training of deep configurations, combining RNN and energy model characteristics.

\section{Experimental Setting}
\subsection{Dataset: BHDD}
\label{subsec:dataset}

To benchmark the performance of the selected models, we utilize the \texttt{BHDD} dataset \cite{aung2026bhdd}, \cite{expaAI2024BHDD}. This dataset is designed to be the direct Burmese-language counterpart to the renowned MNIST dataset, providing a standardized benchmark for evaluating machine learning and deep learning models on the Burmese digit script.

The dataset was curated from a diverse group of over 150 individuals, encompassing a wide range of ages (from high school students to professionals in their 50s) and occupations. This collection strategy ensures a rich variety of handwriting styles, making the dataset robust and representative of real-world variation.

The structure and format of \texttt{BHDD} are intentionally aligned with the original MNIST to ensure ease of use and direct comparability of models. The key statistics of the dataset are as follows:

\begin{itemize}
    \item \textbf{Training Set:} 60,000 grayscale images.
    \item \textbf{Testing Set:} 27,561 grayscale images.
    \item \textbf{Number of Classes:} 10 (corresponding to Burmese digits 0 through 9).
    \item \textbf{Image Dimensions:} Each image is $28 \times 28$ pixels, stored in a flattened format as a 1D array of 784 elements.
    \item \textbf{Labels:} All labels are provided in a one-hot encoded format.
\end{itemize}

\begin{figure}[htbp]
\centerline{\includegraphics[width=\columnwidth, keepaspectratio]{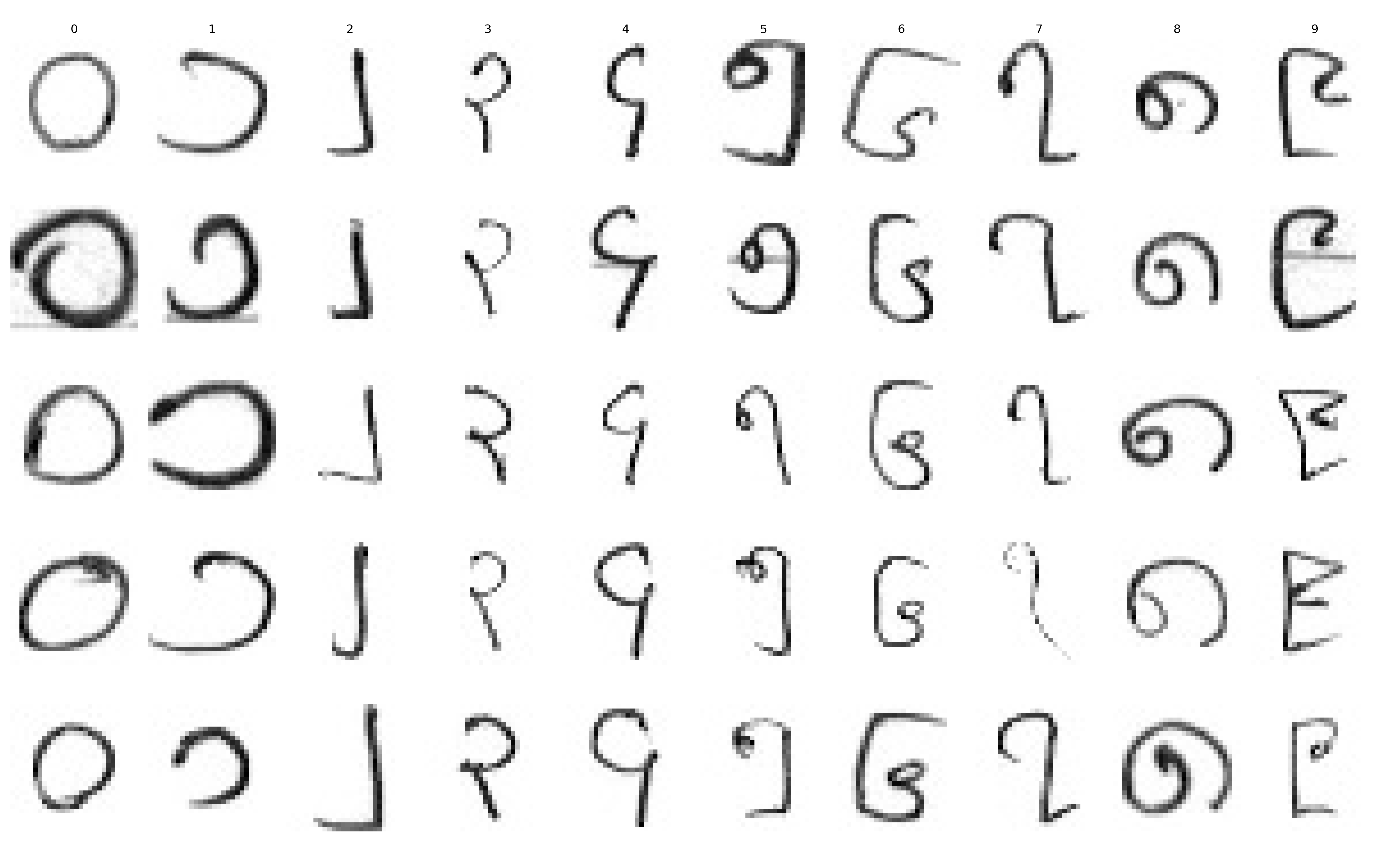}}
\caption{Visualization of the \texttt{BHDD} dataset. The figure displays five randomly selected samples for each of the ten Burmese digit classes (0-9), illustrating the variety of handwriting styles present in the dataset.}
\label{fig:myMNIST_samples}
\end{figure}

\noindent Figure~\ref{fig:myMNIST_samples} provides a visual overview of the dataset, showcasing the variation in handwriting styles for each digit. The dataset is publicly available under the LGPL-3.0 license and is intended to foster innovation and benchmarking within the community for Burmese language AI research \cite{expaAI2024BHDD}.

\subsection{Practical Implementation and Training Details}

All models were implemented in PyTorch~\cite{paszke2019pytorch} and trained on a single NVIDIA GeForce RTX 3090 Ti GPU with 24~GB of VRAM. To ensure a fair and reproducible comparison, a consistent training configuration was applied across all models where applicable.

\subsubsection{Training Configuration}
The standard training hyperparameters were as follows:
\begin{itemize}
\item \textbf{Optimizer:} AdamW~\cite{loshchilov2017decoupled} with a learning rate of $3 \times 10^{-4}$ and weight decay of $10^{-4}$.
\item \textbf{Learning Rate Schedule:} OneCycleLR policy with a maximum learning rate of $5 \times 10^{-4}$.
\item \textbf{Batch Size:} 128 for training and 1,000 for validation and testing.
\item \textbf{Loss Function:} Cross-Entropy loss.
\item \textbf{Regularization:} Employed dropout (rates between 0.1--0.3), layer normalization, and gradient clipping with a maximum norm of 1.0 to ensure training stability.
\item \textbf{Epochs:} Models were trained for 30 to 100 epochs, with early stopping based on validation loss plateauing to prevent overfitting.
\end{itemize}

\subsubsection{Computational Performance and Resource Utilization}
Training times and resource consumption varied significantly based on model architectural complexity, particularly for recurrent and attention-based models. Table~\ref{tab:training_resources} summarizes the computational footprint of each model type for a 50-epoch run.

\begin{table*}[!htbp]
\centering
\begin{tabular}{p{3cm} c c c}
\hline
\textbf{Model Type} & \textbf{Time per Epoch} & \textbf{Total Time} & \textbf{GPU Memory (GB)} \\ 
\hline
MLP & $\sim$45 sec & $\sim$38 min & $\sim$1.2 \\
CNN & $\sim$50 sec & $\sim$42 min & $\sim$1.5 \\
LSTM/GRU & $\sim$90 sec & $\sim$75 min & $\sim$2.8 \\
Transformer & $\sim$110 sec & $\sim$92 min & $\sim$3.5 \\
JEM & $\sim$180 sec & $\sim$150 min & $\sim$4.0 \\
FastKAN/EfficientKAN& $\sim$70 sec & $\sim$58 min & $\sim$2.0 \\
PETNN & $\sim$20 sec & $\sim$17 min & $\sim$1.8 \\ 
\hline
\end{tabular}
\caption{Training Time and GPU Memory Utilization by Model Type (50 Epochs)}
\label{tab:training_resources}
\end{table*}

The PETNN architecture demonstrated notably efficient training, leveraging its gated recurrent mechanism to achieve competitive performance with lower computational overhead compared to other recurrent and attention-based models. All models sustained high GPU utilization ($>$90\%$)$ during training phases.

\subsubsection{Convergence Dynamics and Hyperparameter Sensitivity}
The convergence behavior and sensitivity to hyperparameters were carefully logged:

\begin{itemize}
\item \textbf{MLP/CNN:} These feedforward architectures converged rapidly, typically reaching peak performance within 20--30 epochs. Higher dropout rates ($p=0.25$--$0.3$) were crucial for the CNN to mitigate overfitting.
\item \textbf{RNNs (LSTM/GRU):} Exhibited slower initial convergence, stabilizing after approximately 25 epochs. The GRU variant trained approximately $15\%$ faster than the LSTM while achieving comparable accuracy.
\item \textbf{Transformer:} Required a learning rate warm-up period (5 epochs) for stable training of its self-attention layers. Peak accuracy was consistently observed around epoch 35.
\item \textbf{JEM:} The incorporation of Langevin sampling for contrastive divergence increased per-epoch training time by a factor of 2--3 but yielded better calibrated uncertainty estimates.
\item \textbf{KANs:} Exhibited a very smooth and stable loss descent. The EfficientKAN implementation trained $\sim$20\% faster than FastKAN due to its optimized spline activation functions.
\item \textbf{PETNN:} Performance was sensitive to the gating activation function. The SiLU activation gate achieved the best results, while sigmoid gates occasionally led to saturation issues. Optimal hidden and cell dimensions were found to be 192 and 48, respectively.
\end{itemize}

\subsubsection{Reproducibility}
All experiments were conducted with fixed random seeds (42) to ensure reproducibility. The complete source code, including all model implementations and training scripts, is publicly available for review and replication at: \url{https://github.com/ye-kyaw-thu/myMNIST-benchmark}. A summary of the key architectural specifications for all baseline models is provided in Table~\ref{tab:model_specs}.

\begin{table}[htbp]
\centering
\begin{tabular}{lcc}
\hline
\textbf{Model} & \textbf{Layers} & \textbf{Hidden Dim} \\
\hline
MLP & 3 & [256, 128] \\
CNN & 4 & [32, 64] \\
LSTM & 2 & 192 \\
GRU & 2 & 192 \\
Transformer & 2 & 64 \\
JEM & 5 & [16, 32, 64] \\
FastKAN & 2 & 64 \\
EfficientKAN & 2 & 64 \\
PETNN-Sigmoid & 3 & 192 \\
PETNN-GELU & 3 & 192\\
PETNN-SiLU & 3 & 192\\
\hline
\end{tabular}
\caption{Model architecture specifications}
\label{tab:model_specs}
\end{table}

\section{Results and Discussion}
\label{sec:results}
The performance of all eleven models on the myMNIST benchmark is quantitatively summarized in Table~\ref{tab:results}, with key metrics including Precision, Recall, F1-Score, and Accuracy. The results reveal clear hierarchies in model efficacy for the task of Burmese handwritten digit recognition.

\begin{table*}[t!]
\centering
\begin{tabular}{lcccc}
\hline
\textbf{Model} & \textbf{Precision} & \textbf{Recall} & \textbf{F1-Score} & \textbf{Accuracy} \\ \hline
MLP            & 0.9810             & 0.9895          & 0.9852            & 0.9907            \\
CNN            & 0.9955             & 0.9963          & \textbf{0.9959}   & \textbf{0.9970}   \\
LSTM           & 0.9907             & 0.9942          & 0.9924            & 0.9951            \\
GRU            & 0.9886             & 0.9934          & 0.9910            & 0.9937            \\
Transformer    & 0.9898             & 0.9944          & 0.9921            & 0.9946            \\
JEM            & 0.9931             & 0.9957          & 0.9944            & 0.9958            \\
FastKAN        & 0.9844             & 0.9914          & 0.9879            & 0.9922            \\
EfficientKAN   & 0.9841             & 0.9898          & 0.9869            & 0.9918            \\
PETNN (Sigmoid) & 0.9897            & 0.9940          & 0.9918            & 0.9943            \\
PETNN (GELU)   & 0.9947             & 0.9963          & 0.9955            & 0.9966            \\
PETNN (SiLU)   & 0.9944             & 0.9961          & 0.9952            & 0.9964            \\ \hline
\end{tabular}
\caption{Performance comparison of various models on the BHDD dataset.}
\label{tab:results}
\end{table*}

\subsection{Overall Performance Hierarchy}

The Convolutional Neural Network (CNN) emerged as the top-performing architecture, achieving the highest scores in both F1-Score (0.9959) and Accuracy (0.9970). This reaffirms the well-established effectiveness of convolutional inductive biases for image-related tasks, as the CNN excels at capturing spatial hierarchies and local patterns inherent in digit images. Notably, the physics-inspired PETNN variants demonstrated highly competitive performance. The PETNN (GELU) model was a close second, nearly matching the CNN's performance (F1: 0.9955, Accuracy: 0.9966), while PETNN (SiLU) also performed strongly. This indicates that the energy transition mechanism in PETNN provides a powerful alternative to classical convolutional feature extraction.

The Joint Energy Model (JEM) performed commendably, securing a position among the top tiers (F1: 0.9944, Accuracy: 0.9958). Its success suggests that hybrid generative-discriminative learning can be a viable approach for this domain. Among recurrent and attention-based models, the results followed an expected pattern: the LSTM (F1: 0.9924) slightly outperformed the Transformer (0.9921) and GRU (0.9910), though all were surpassed by the CNN and top PETNN variants. 



The Multi-Layer Perceptron (MLP) served as a lower baseline, as anticipated, due to its lack of architectural priors for processing image data. The recently proposed Kolmogorov-Arnold Networks (KANs), specifically FastKAN and EfficientKAN, delivered respectable but not state-of-the-art results (Accuracy $\approx$ 0.992), positioning them as meaningful yet less effective alternatives to the leading models in this benchmark.

\subsection{Analysis of PETNN Variants}

The performance across the three PETNN activation functions offers insights into its dynamics. The \textbf{GELU} activation yielded the best results, likely due to its smooth, non-monotonic nature which can better approximate complex functions and improve gradient flow compared to the sigmoidal gating commonly used in LSTMs \cite{hendrycks2016gaussian}. The \textbf{SiLU} activation performed similarly well, supporting this hypothesis. The standard \textbf{Sigmoid} activation, while effective, trailed its counterparts, suggesting that the choice of activation function is crucial for unlocking the full potential of the PETNN's physics-inspired memory mechanism.

\section{Error Analysis}

\begin{figure*}[htbp]
\centering
\begin{subfigure}[b]{0.32\textwidth}
\centering
\fbox{\includegraphics[width=\linewidth]{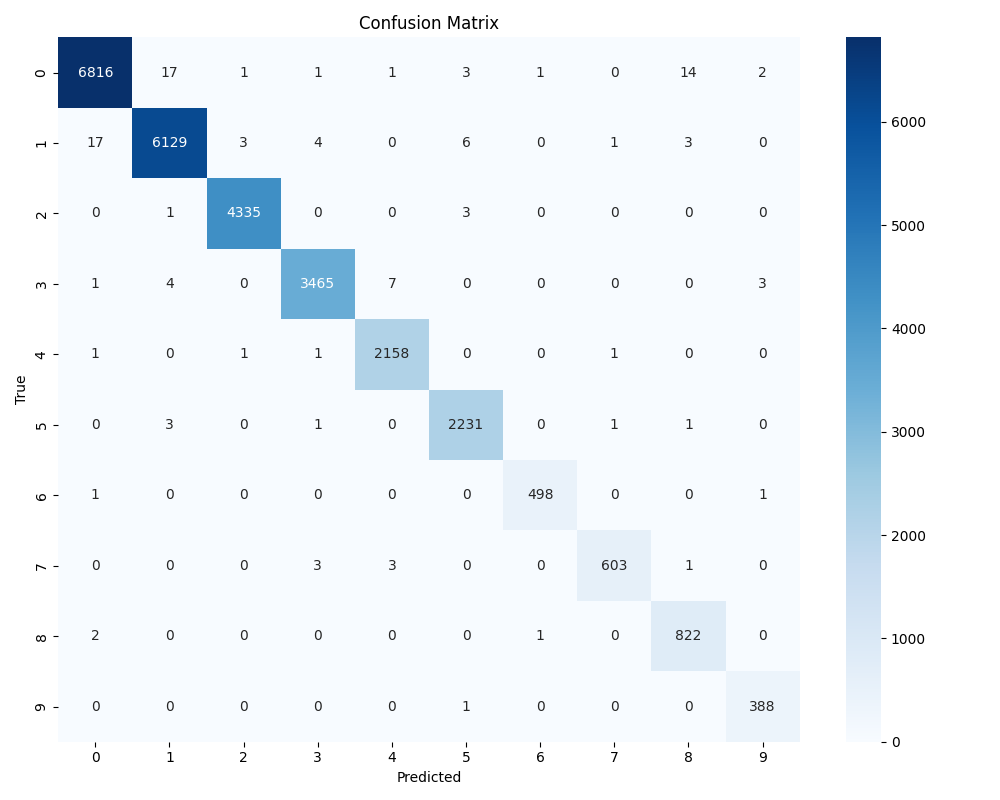}}
\caption{JEM.}
\label{fig:cm_JEM}
\end{subfigure}
\hfill
\centering
\begin{subfigure}[b]{0.32\textwidth}
\centering
\fbox{\includegraphics[width=\linewidth]{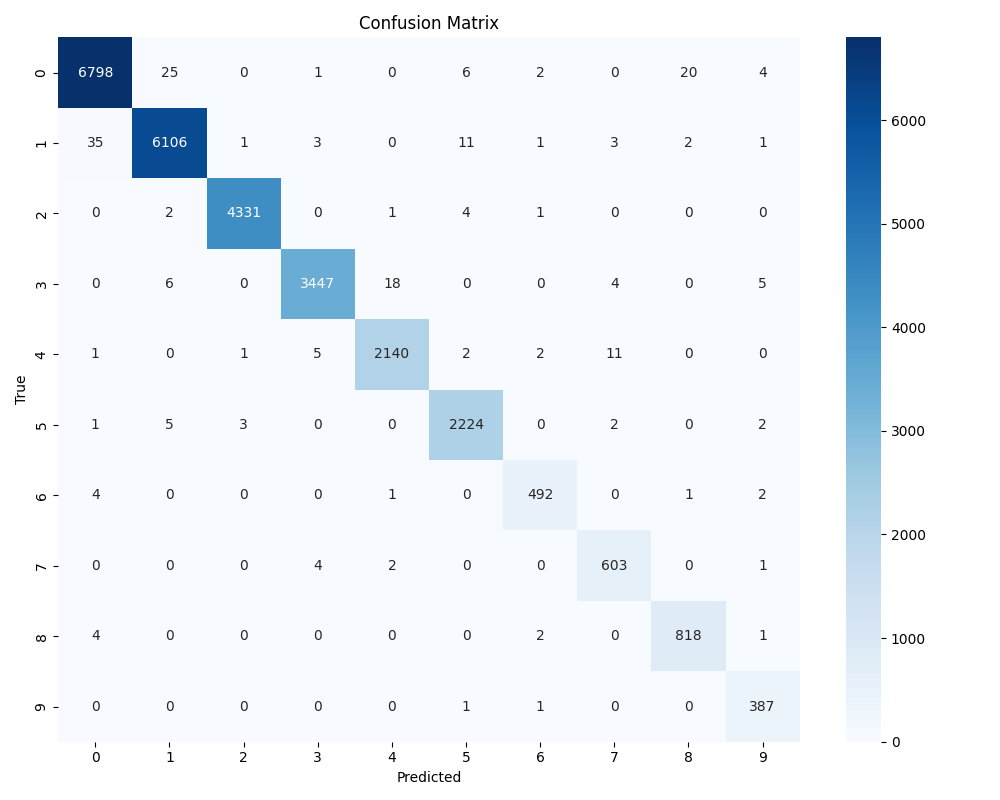}}
\caption{fastKAN.}
\label{fig:cm_fastkan}
\end{subfigure}
\hfill
\begin{subfigure}[b]{0.32\textwidth}
\centering
\fbox{\includegraphics[width=\linewidth]{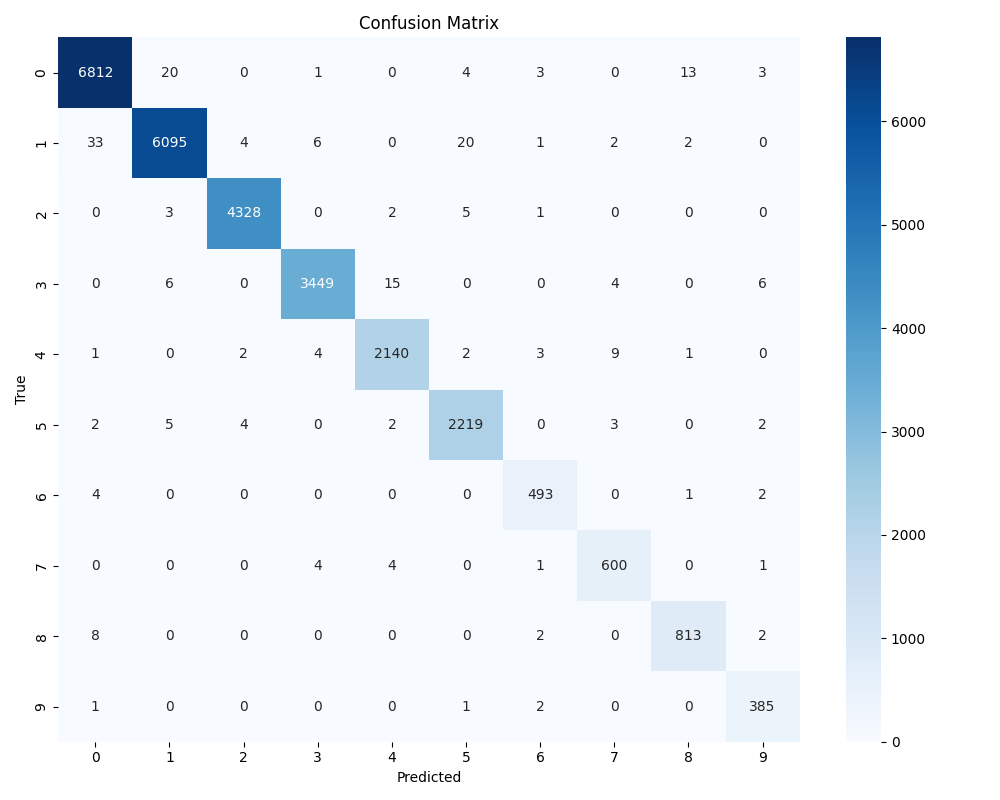}}
\caption{efficientKAN.}
\label{fig:cm_effikan}
\end{subfigure}
\hfill
\begin{subfigure}[b]{0.32\textwidth}
\centering
\fbox{\includegraphics[width=\linewidth]{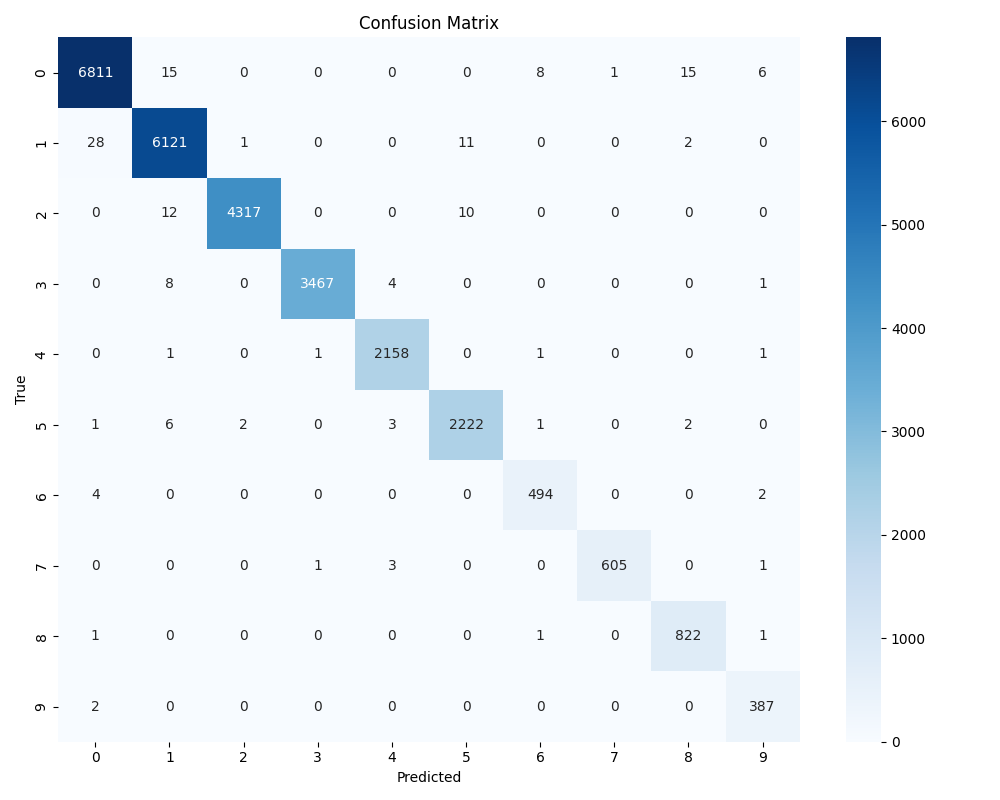}}
\caption{PETNN (Sigmoid).}
\label{fig:cm_petnn_sigmoid}
\end{subfigure}
\hfill
\begin{subfigure}[b]{0.32\textwidth}
\centering
\fbox{\includegraphics[width=\linewidth]{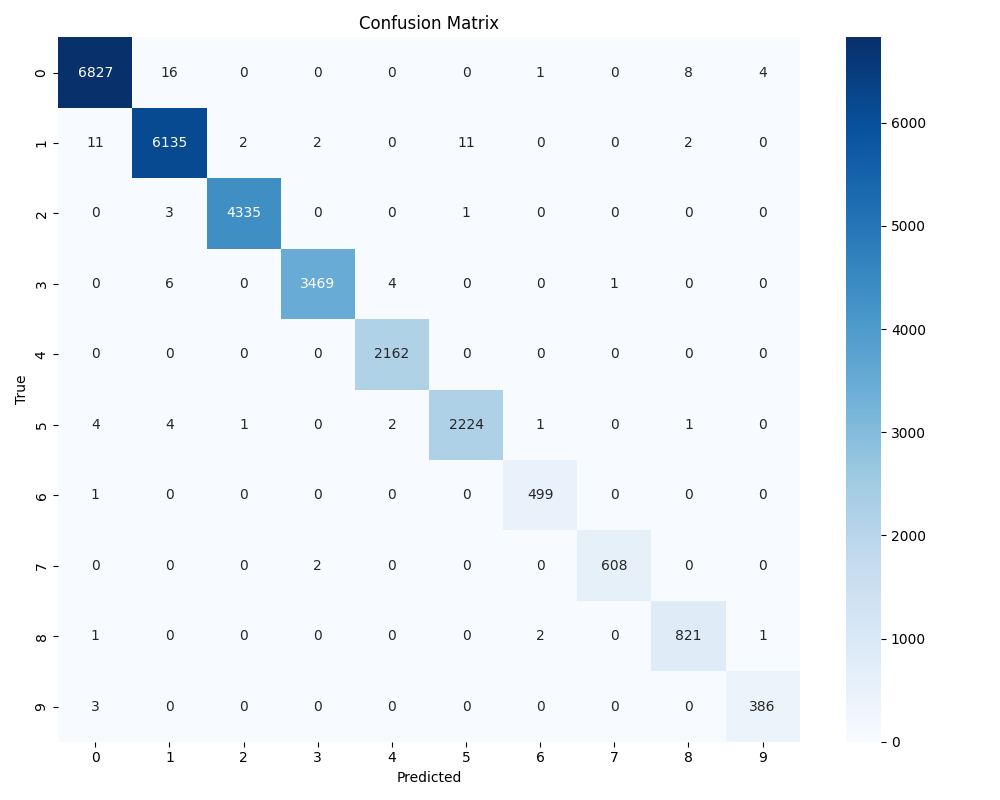}}
\caption{PETNN (GELU).}
\label{fig:cm_petnn_gelu}
\end{subfigure}
\hfill
\begin{subfigure}[b]{0.32\textwidth}
\centering
\fbox{\includegraphics[width=\linewidth]{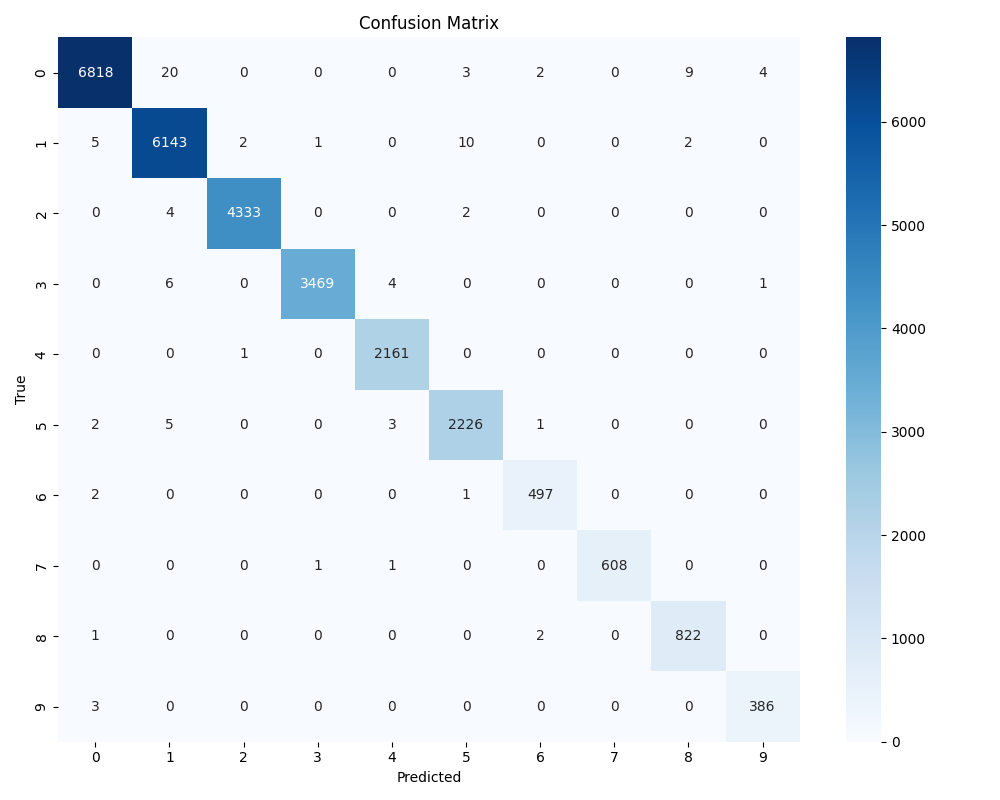}}
\caption{PETNN (SiLU).}
\label{fig:cm_petnn_silu}
\end{subfigure}
\caption{Confusion matrices for the evaluated models, highlighting common misclassification patterns.}
\label{fig:confusion_matrices}
\end{figure*}

The performance of all evaluated architectures was analyzed using the confusion matrices illustrated in Figure \ref{fig:confusion_matrices}. The models exhibit generally high accuracy, with the PETNN (GELU) and JEM architectures achieving the strongest overall results (Accuracy $\approx 0.996\%$). A clear performance hierarchy exists among the PETNN variants, demonstrating that model robustness is highly sensitive to the activation function: PETNN (GELU) and PETNN (SiLU) (Accuracy $0.9966$ and $0.9964$) significantly outperformed PETNN-Sigmoid (Accuracy $ 0.9943$). Similarly, the KAN-based models, fastKAN and efficientKAN (both $\approx 0.992\%$), showed slightly lower accuracy and a higher overall volume of classification errors compared to the top-performing PETNN and JEM models.

A detailed examination of the off-diagonal elements reveals that recognition errors are highly localized, concentrated primarily around the true digits 1 (\textbf{\padauktext{၁}}) and 0 (\textbf{\padauktext{၀}}). Across almost all models, the top three misclassification pairs involve these two classes. For instance, the 1 (\textbf{\padauktext{၁}}) $\rightarrow$ 0 (\textbf{\padauktext{၀}}) error is the single most frequent mistake for the fastKAN (35 times), efficientKAN (33 times), and PETNN-Sigmoid (28 times). Conversely, $0 \rightarrow 1$ is the largest error for PETNN (GELU), PETNN (SiLU), and JEM, confirming a fundamental bi-directional confusion between these two glyphs. The true digit 1 (\textbf{\padauktext{၁}}), in particular, proves to be the most challenging class overall, as it is also frequently misclassified as 5 (\textbf{\padauktext{၅}}), especially by the PETNN models.

These prevalent confusions stem directly from the visual similarity and inherent stroke ambiguity in the handwritten Myanmar script.

\begin{itemize}
    \item 1 (\textbf{\padauktext{၁}}) $\leftrightarrow$ 0 (\textbf{\padauktext{၀}}): The glyph 1 (\textbf{\padauktext{၁}}) is defined as an open, near−circular stroke that does not fully enclose (i.e., it has a break or gap on the left side), whereas 0 (\textbf{\padauktext{၀}}) is a simple circle. The difficulty in distinguishing them arises in handwritten contexts: If the 1(\textbf{\padauktext{၁}}) is written quickly, the stroke may unintentionally close completely, causing it to collapse visually and be misclassified as a 0(\textbf{\padauktext{၀}}). Conversely, a sloppy or noisy handwritten 0(\textbf{\padauktext{၀}}) might contain a tiny gap or a slight vertical bulge, leading the model to perceive it as the 1(\textbf{\padauktext{၁}})'s open structure and stem. 
    
    \item 1 (\textbf{\padauktext{၁}}) $\leftrightarrow$ 5 (\textbf{\padauktext{၅}}): The 5 (\textbf{\padauktext{၅}}) glyph, a circle with a longer tail, is also confused with 1 when its tail is written too tightly or truncated. 
    
    
    \item Directional Confusion 3 (\textbf{\padauktext{၃}}) $\rightarrow$ 4 (\textbf{\padauktext{၄}}): This directional error is a key weakness in the KAN architectures (fastKAN: 18, efficientKAN: 15). The confusion stems from the very subtle stroke directionality that distinguishes these two glyphs. The digit 3 (\textbf{\padauktext{၃}}) is fundamentally an open, near-circular form with the opening on the left, whereas 4 (\textbf{\padauktext{၄}}) is an open, near-circular form with the opening on the right. The misclassification 3 $\rightarrow$ 4 likely occurs when the model fails to correctly distinguish the orientation of the open side or misinterprets the start/end point of the single primary stroke, leading to a confusion of the 3's left-open structure for the 4's right-open structure. This indicates the KAN models may struggle with fine-grained feature extraction of stroke direction and closure position.
    
\end{itemize}


\section{Conclusion and Future Work}
\label{sec:conclusion}

This study presented a comprehensive benchmark of eleven diverse neural architectures on the BHDD dataset for Burmese handwritten digit recognition. Our results provide a clear performance hierarchy, establishing robust baselines for future research in this domain. The Convolutional Neural Network (CNN) confirmed its status as a powerful baseline for image classification, while the physics-inspired PETNN model, particularly its GELU variant, demonstrated remarkable competitiveness, validating its novel design principle. The strong performance of the Joint Energy Model (JEM) and the respectable results from Kolmogorov-Arnold Networks (KANs) further enrich the landscape of viable approaches for this task.


\begin{thebibliography}{00}

\bibitem{aung2026bhdd} S. H. Aung, H. Htet, H. S. W. Khaing, and T. M. Nyunt, ``BHDD: A Burmese Handwritten Digit Dataset,'' arXiv preprint arXiv:2603.21966, 2026. [Online]. Available: \url{https://arxiv.org/abs/2603.21966}

\bibitem{expaAI2024BHDD} Expa.AI Research Team, ``Burmese Handwritten Digit Dataset (BHDD),'' Online, 2024. [Online]. Available: \url{https://github.com/baseresearch/BHDD} [Accessed: 2025-05-10]

\bibitem{ThuraAungBHDD} T. Aung, ``Convolutional Neural Network Training Notebook for BHDD,'' Online, 2024. [Online]. Available: \url{https://colab.research.google.com/github/ThuraAung1601/BHDD-using-streamlit/blob/main/CNN\_train.ipynb}

\bibitem{LeCun1998GradientbasedLA} Y. LeCun, L. Bottou, Y. Bengio, and P. Haffner, ``Gradient-based learning applied to document recognition,'' Proc. IEEE, vol. 86, no. 11, pp. 2278--2324, Nov. 1998.

\bibitem{Ahmed2023NovelTechnique} S. S. Ahmed, Z. Mehmood, I. A. Awan, and R. M. Yousaf, ``A Novel Technique for Handwritten Digit Recognition Using Deep Learning,'' J. Sensors, vol. 2023, p. 2753941, 2023. doi: 10.1155/2023/2753941.

\bibitem{efficientKANGitHub} B. Li, ``Efficient-KAN Repository,'' Online, 2024. [Online]. Available: \url{https://github.com/Blealtan/efficient-kan}

\bibitem{fastKANGitHub} Z. Li, ``Fast-KAN Repository,'' Online, 2024. [Online]. Available: \url{https://github.com/ZiyaoLi/fast-kan}

\bibitem{glorot2010understanding} X. Glorot and Y. Bengio, ``Understanding the difficulty of training deep feedforward neural networks,'' in Proc. 13th Int. Conf. Artif. Intell. Statist., 2010, pp. 249--256.

\bibitem{loshchilov2017decoupled} I. Loshchilov and F. Hutter, ``Decoupled weight decay regularization,'' arXiv preprint arXiv:1711.05101, 2017.

\bibitem{hornik1989multilayer} K. Hornik, M. Stinchcombe, and H. White, ``Multilayer feedforward networks are universal approximators,'' Neural Netw., vol. 2, no. 5, pp. 359--366, 1989.

\bibitem{simonyan2014very} K. Simonyan and A. Zisserman, ``Very deep convolutional networks for large-scale image recognition,'' arXiv preprint arXiv:1409.1556, 2014.

\bibitem{he2015delving} K. He, X. Zhang, S. Ren, and J. Sun, ``Delving deep into rectifiers: Surpassing human-level performance on imagenet classification,'' in Proc. IEEE Int. Conf. Comput. Vis. (ICCV), 2015, pp. 1026--1034.

\bibitem{zeiler2014visualizing} M. D. Zeiler and R. Fergus, ``Visualizing and understanding convolutional networks,'' in Comput. Vis.--ECCV, D. Fleet, T. Pajdla, B. Schiele, and T. Tuytelaars, Eds. Springer, 2014, pp. 818--833.

\bibitem{saxe2013exact} A. M. Saxe, J. L. McClelland, and S. Ganguli, ``Exact solutions to the nonlinear dynamics of learning in deep linear neural networks,'' arXiv preprint arXiv:1312.6120, 2013.

\bibitem{strobelt2016lstm} H. Strobelt, S. Gehrmann, M. Behrisch, A. Perer, H. Pfister, and A. M. Rush, ``LSTMVis: A tool for visual analysis of hidden state dynamics in recurrent neural networks,'' IEEE Trans. Vis. Comput. Graphics, vol. 24, no. 1, pp. 667--676, 2016.

\bibitem{grathwohl2019your} W. Grathwohl, K. Wang, J. Jacobsen, D. Duvenaud, M. Norouzi, and K. Swersky, ``Your classifier is secretly an energy based model and you should treat it like one,'' arXiv preprint arXiv:1912.03263, 2019.

\bibitem{li2024kan} Z. Liu, Y. Wang, S. Vaidya, F. Ruehle, J. Halverson, M. Solja\v{c}i\'{c}, T. Y. Hou, and M. Tegmark, ``KAN: Kolmogorov-Arnold Networks,'' arXiv preprint arXiv:2404.19756, 2024.

\bibitem{wu2025physicsinspiredenergytransitionneural} Z. Wu, J. An, B. Xu, F. Shen, and J. Zhao, ``Physics-inspired Energy Transition Neural Network for Sequence Learning,'' arXiv preprint arXiv:2505.03281, 2025.

\bibitem{petnncolab} PETNN Research Group, ``Physics-inspired Energy Transition Neural Network (PETNN) Implementation Demo,'' Online, 2024. [Online]. Available: \url{https://colab.research.google.com/drive/1GOnxXHP8sqMMn4AeIA2P8Cl9ozbRwXqc?usp=sharing}

\bibitem{hendrycks2016gaussian} D. Hendrycks and K. Gimpel, ``Gaussian Error Linear Units (GELUs),'' arXiv preprint arXiv:1606.08415, 2016.

\bibitem{elfwing2017sigmoid} S. Elfwing, E. Uchibe, and D. Doya, ``Sigmoid-weighted linear units for neural network function approximation in reinforcement learning,'' Neural Netw., vol. 107, pp. 3--11, 2017.

\bibitem{paszke2019pytorch} A. Paszke et al., ``PyTorch: An imperative style, high-performance deep learning library,'' in Adv. Neural Inf. Process. Syst., 2019.

\end{thebibliography}
\end{document}